\documentclass{article}
\usepackage[T1]{fontenc}
\usepackage{spconf,amsmath,amssymb,graphicx,booktabs}
\usepackage[hidelinks]{hyperref}
\title{SEQUENTIAL AUDITING OF CONNECTED FORECAST ERRORS UNDER PARTIAL OBSERVATIONS}
\name{Yuanbo Zhou \qquad Junjie Xiong}
\address{Missouri University of Science and Technology\\
\texttt{yztpn@mst.edu}\qquad\texttt{junjiexiong@mst.edu}}
\begin{document}
\raggedbottom
\maketitle
\begin{abstract}
Graph forecasting systems often rely on prediction intervals to quantify uncertainty across connected locations, but the true outcomes needed for evaluation may be only partially observed. We study how an auditor can selectively retrieve missing outcomes to determine whether there exists a connected group whose outcomes violate the same side of their respective forecast intervals. The audit stops when the observed labels determine the event or the query budget is exhausted. On three public graph time series, MOMS reduces mean queries by 19.6\% to 38.5\% relative to screened random acquisition when 30\% of labels are hidden. Under localized masking on METR-LA, shortest witness finds more positive cases within ten queries, although the difference depends on node ordering. Direct component traversal reproduces all 9,696 screened random traces without enumerating connected subsets. The experiments show that reducing queries to a decision does not necessarily improve early event discovery or reduce local computation.
\end{abstract}

\section{Introduction}
Graph forecasting uses relations among locations to predict network
measurements~\cite{shuman2013emerging,li2018dcrnn}. Prediction intervals quantify uncertainty in individual outcomes~\cite{romano2019cqr,xu2021enbpi}. When some outcomes are unavailable to an auditor, checking a regional forecast error requires deciding which measurements to retrieve. We ask whether this decision can be made before all missing labels are collected.

The event of interest is a connected group of at least $k$ nodes whose outcomes fall below their respective lower interval endpoints or above their upper endpoints. Retrieved labels are assumed exact, and permanently lost measurements are excluded. These interval violations define the audit target rather than verified physical faults. Graph sampling instead studies signal recovery under assumptions such as bandlimitedness~\cite{chen2015sampling}.
Predictive monitoring evaluates requirements over uncertain future
trajectories~\cite{ma2021predictive}. Our audit uses realized outcomes and keeps the forecasts fixed during acquisition.

Graph anomaly detection uses scan statistics and connectivity to identify regions with unusual observations~\cite{sharpnack2013less,cadena2018}. IISD studies interactive detection of connected subgraphs~\cite{yu2022iisd}, while MITIGATE acquires labels to train graph anomaly detectors~\cite{chang2024mitigate}. Here, labels are retrieved to decide a specified event. Confirming one region proves presence, whereas absence requires ruling out all compatible regions.

This decision problem is related to stochastic Boolean function and DNF evaluation~\cite{deshpande2013sbfe,allen2014dnf}. Adaptive submodularity and Bayesian active learning also study observation selection under uncertainty~\cite{golovin2011adaptive,golovin2010noisy}. Our formulation uses categorical labels with mutually exclusive lower and upper tails. We derive an exact stopping condition and compare adaptations of MOMS and JW with other acquisition rules~\cite{mahfoudh2003,jeroslow1990}.

Experiments use three public graph time series~\cite{li2018dcrnn,rozemberczki2021pgt}, fixed prediction intervals and simulated visibility masks. We measure queries to a decision, positive discoveries within a budget and computation time. The comparisons examine when these criteria favor different policies and whether the rankings persist after node reindexing. The contribution is the audit formulation and its empirical characterization using established selection principles.

\begin{figure*}[t]
\centering
\includegraphics[width=\textwidth]{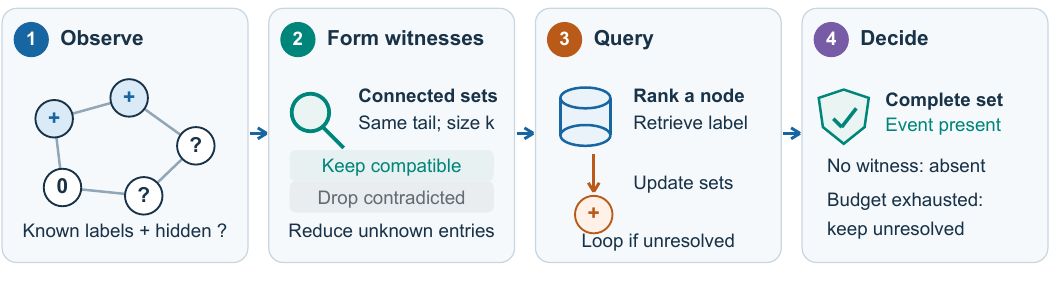}
\caption{Illustrative certificate audit. Symbols distinguish known interval labels ($+$, $0$) from unknowns ($?$). Retrieved labels update compatible witnesses until presence or absence is established, or the budget is exhausted.}
\label{fig:method}
\end{figure*}

\section{Audit Task and Exact Decisions}
At a fixed time, let $G=(V,E)$ be the undirected graph of eligible nodes. Node $i$ has outcome $y_i$, forecast interval $[L_i,U_i]$ and label
\begin{equation}
z_i=\begin{cases}
-1,&y_i<L_i,\\
0,&L_i\leq y_i\leq U_i,\\
+1,&y_i>U_i.
\end{cases}
\end{equation}
Let $\mathcal C_k$ contain all connected subsets of $k$ nodes. Define
\begin{equation}
F_k(z)=\mathbf1\!\left\{\exists C\in\mathcal C_k,\ s\in\{-1,+1\}:
z_i=s\ \forall i\in C\right\}.
\label{eq:event}
\end{equation}
Every connected component of at least $k$ equally labeled nodes contains such a subset. The auditor initially observes a partial label assignment. Each query reveals one hidden eligible label at unit cost. Neither the graph nor the intervals change during the audit.

A pair $(C,s)$ is a positive witness. An observed label that differs from $s$ invalidates it. Matching labels leave only its unknown entries to be queried. The residual family $\mathcal R$ records these entries and the sign of each surviving witness (Fig.~\ref{fig:method}). A completed witness certifies presence. If no witness survives, the event is absent. The audit otherwise remains unresolved and may continue within its query budget.

The stopping rule is exact under truthful retrieval. If an incomplete witness survives and none is complete, assigning its unknown labels the matching sign produces a positive completion. Assigning zero to every
unknown label produces a negative completion. Both outcomes are therefore
consistent with the observations, so an exact decision cannot yet be made.
This argument applies to~\eqref{eq:event} on eligible nodes.

The same decision can be checked through graph traversal. For each sign, consider the subgraphs induced by known matching nodes and by matching or unknown nodes. A known component of size at least $k$ certifies presence. If neither compatible subgraph contains such a component, the event is absent. Every unknown node in a compatible component of size at least $k$ belongs to a possible witness.  Screened random acquisition can therefore identify relevant nodes without enumerating connected subsets. Our implementation stores adjacency as Python integer bitsets.

\begin{table*}[t]
\centering
\caption{Mean queries among initially unresolved audits at 30\% masking. All policies use the same exact stopping rule. Lower values indicate fewer queries.}
\label{tab:main}
\begin{tabular}{llrrrrr}
\toprule
Dataset & Mask & Screened random & MOMS & Shortest witness & Shortest pool & JW\\
\midrule
WikiMaths64 & Random & 3.389 & 2.287 & 2.563 & 2.289 & 2.305\\
 & Localized & 12.977 & 7.974 & 8.650 & 7.988 & 8.014\\
METR-LA & Random & 35.997 & 25.761 & 29.252 & 25.856 & 26.174\\
 & Localized & 51.763 & 41.618 & 44.401 & 41.756 & 42.035\\
MontevideoBus & Random & 18.019 & 14.432 & 18.535 & 14.408 & 14.587\\
 & Localized & 102.310 & 67.511 & 150.621 & 67.470 & 67.874\\
\bottomrule
\end{tabular}
\end{table*}

\section{Query Policies and Implementation}
All policies apply the stopping rule before querying and after each returned label. Fixed ID follows node indices. Fixed degree orders hidden nodes by degree in the eligible induced graph. Random ordering draws a permutation without using hidden outcomes. Screened random follows that permutation but skips nodes absent from every surviving witness. Comparing these two random policies isolates the effect of removing irrelevant queries.

MOMS prioritizes variables with maximum occurrences in clauses of minimum size~\cite{mahfoudh2003}. In our adaptation, residual sets are deduplicated within each sign. If $d$ is the smallest residual size, the next node maximizes
\begin{equation}
S_i=\sum_{(R,s)\in\mathcal R}\mathbf1\{|R|=d,\ i\in R\}.
\label{eq:moms}
\end{equation}
The JW adaptation sums $2^{-|R|}$ over the signed residual sets containing
node $i$~\cite{mahfoudh2003,jeroslow1990}. Both rules select which label to retrieve. Recorded outcomes supply the returned labels.

Shortest witness selects the first surviving witness with the fewest
unknown nodes and queries its unknown node of highest degree. Shortest pool restricts candidates to witnesses of minimum residual size, then counts their occurrences across all surviving original witnesses. Raw count uses the same counts without restricting the candidates. Ties follow node index order. These controls retain original graph degrees in the main comparison. Only fixed degree uses the degrees of the eligible induced graph.

The reference MOMS code deduplicates signed residual sets using Python tuples. The packed implementation encodes sorted indices and signs as unsigned integers, checks for overflow and deduplicates the resulting keys. It preserves both scores and tie breaking. Verification checks that both implementations return identical query sequences and decisions.

\begin{figure*}[t]
\centering
\includegraphics[width=\textwidth]{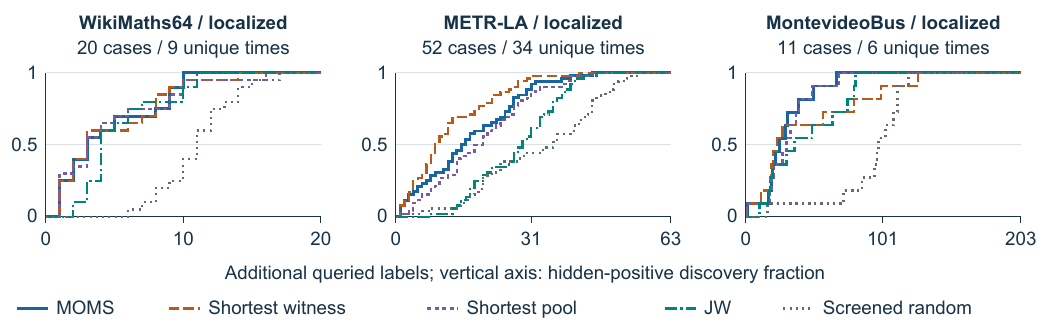}
\caption{Discovery of hidden connected triples across query budgets under 30\% localized masking. Each panel reports positive cases and distinct positive timestamps. Line styles distinguish policies in grayscale. Queries reveal actual withheld labels. Repeated masks do not create independent events.}
\label{fig:discovery}
\end{figure*}

\section{Experimental Protocol}
\subsection{Data and forecast intervals}
We use METR-LA~\cite{li2018dcrnn} and WikiMaths and MontevideoBus from PyTorch Geometric Temporal~\cite{rozemberczki2021pgt}. The WikiMaths subset, selected with seed 2026, has 64 nodes, 77 edges and 147 evaluation days. METR-LA has 207 nodes and 1,313 edges before eligibility filtering, with 512 evaluation hours. MontevideoBus has 675 nodes, 690 edges and 149 evaluation hours. For METR-LA, hourly aggregation requires at least nine valid readings among the twelve measurements taken at five-minute intervals. Zero and nonfinite raw speeds are treated as missing. Missing targets are excluded from each audit graph, and 60 entirely missing hours are omitted. Zero counts in MontevideoBus are valid targets.

A ridge predictor uses four lags of each node, four lags of the aggregated neighbor signal and an intercept. Its parameters are shared across nodes. The first 60\% of timestamps fit the predictor, the next 10\% fit residual quantiles and evaluation starts at 80\%. Scaling uses the corresponding fitting block. Missing METR-LA inputs are filled from preceding observations, but filled targets are never used as ground truth. Quantiles at 0.05 and
0.95 use eight features derived from signed residuals and their magnitudes. The fits use 4,672 rows for WikiMaths64 and fixed samples of 8,192 rows for each larger dataset. Residual quantile regression supplies the initial endpoints~\cite{romano2019cqr}. We expand normalized intervals to contain both pooled marginal and uniform connected triple corrections. The corrections use 52 preceding timestamps for WikiMaths and MontevideoBus and 168 hours for METR-LA. All policies use the same saved intervals.

\subsection{Acquisition and evaluation}
We hide 10\% or 30\% of eligible labels, rounding the count upward. Random masks sample uniformly. Localized masks grow by breadth first traversal and restart in another eligible component when needed. Mask seeds are 73, 101 and 131. Policies share the initial observations, and queries reveal the actual withheld labels. Earlier outcomes remain available for constructing the intervals used at later timestamps.

The primary event size is $k=3$. We report mean queries among initially unresolved audits, resolution at each integer budget and discovery among initially unresolved positive cases. Positive cases that remain unresolved count as undiscovered. Repeated masks at one timestamp are dependent, so we also report the number of distinct positive timestamps. Paired intervals use 3,000 moving block bootstrap replicates, with blocks of 14 days for WikiMaths and 24 hours otherwise. All mask seeds at a timestamp remain grouped. These intervals describe data already inspected during development.

The comparison includes nine policies on 9,696 shared cases, giving 87,264 traces. On an Apple M4 CPU, a fixed hash selects 24 initially unresolved cases per dataset, masking rate and mask type. Three interleaved timing rounds include certificate initialization and updates. Loading, graph preprocessing and label retrieval are excluded. For each round, we average the case times. The reported timing is the median of these three averages.

Additional checks at 30\% masking use three node permutations with seeds 211, 307 and 401. They preserve the physical graph, masks and outcomes. These checks use eligible induced degrees wherever a policy uses degree, and candidate sets are sorted after permutation. Query means use 96 initially unresolved cases selected by hash per condition. Discovery is evaluated on all hidden positive cases, without adding them to the sample for query means. The computation comparison uses the same 24 timing cases and three rounds for enumerated screening, component screening and packed MOMS.

\begin{table}[t]
\centering
\caption{Audit computation in ms: 24 shared cases, 30\% masking. R/L: random/localized. E/C: enumerated/component screening. Values are median round means.}
\label{tab:review}
\begin{tabular}{lrrr}
\toprule
Dataset / mask & E & C & MOMS\\
\midrule
WikiMaths64 / R & 0.052 & 0.071 & 0.070\\
WikiMaths64 / L & 0.141 & 0.159 & 0.243\\
METR-LA / R & 0.583 & 0.823 & 1.030\\
METR-LA / L & 2.342 & 0.976 & 4.939\\
MontevideoBus / R & 0.263 & 1.919 & 0.366\\
MontevideoBus / L & 1.212 & 8.114 & 2.510\\
\bottomrule
\end{tabular}
\end{table}

\section{Results}
\subsection{Queries to a decision}
At 30\% masking, MOMS uses 19.6\% to 38.5\% fewer queries than screened random acquisition (Table~\ref{tab:main}). Across both masking rates, all twelve paired intervals for the query difference exclude zero in favor of MOMS. On METR-LA with random masks, screening alone reduces the mean from 50.37 to 36.00 queries. MOMS further reduces it to 25.76 queries. 

Differences from shortest pool are smaller and vary across datasets.
Shortest pool uses fewer queries than MOMS in all four MontevideoBus
conditions. With 30\% localized masking, their means are 67.470 and 67.511. The reduction relative to random acquisition therefore does not establish a consistent advantage over the other rules that use witness structure.

\subsection{Discovery within ten queries}
Under 30\% localized masking on METR-LA, shortest witness discovers 29 of 52 hidden positive cases within ten queries (Fig.~\ref{fig:discovery}). MOMS discovers 16, shortest pool 14, JW one and screened random three. These 52 cases span 34 distinct hours. Shortest witness requires more queries on average to resolve initially uncertain audits, but discovers more positive cases at this budget in the original node ordering.

In a post hoc check, every true positive triple is entirely hidden in 39 of these 52 cases, compared with five under random masking. In those cases, no revealed label belongs to a true positive triple. A logistic model fitted to preceding history supplies risk scores for a hybrid acquisition rule. The rule uses MOMS when a surviving witness has one unknown label and otherwise selects by risk score. Relative to MOMS, mean queries under localized masking increase from 41.62 to 52.80, while discoveries within ten queries change from 16 to 17. Under random masking, discoveries fall from 47 to 39 of 52. The model is fitted with only 65 positive training labels.

\subsection{Sensitivity to node ordering}
Across the three node permutations, METR-LA localized discovery counts range from 14 to 18 for MOMS and from 17 to 21 for shortest witness. Shortest witness remains ahead in each paired permutation, but the difference is smaller than the original counts of 16 and 29 suggest. Changing only the degree definition leaves those original counts unchanged. For the query means, the MOMS mean minus the shortest pool mean ranges from $-0.219$ to $-0.135$ under localized METR-LA masks. On WikiMaths localized masks, the range is $-0.073$ to $0.021$, so reindexing can reverse the ranking. These ranges summarize the tested permutations and are not confidence intervals for performance on new data.

\subsection{Computation and trace agreement}
Packed MOMS reproduces 9,696 reference traces, including 169,041 query selections. Component screening matches all 9,696 screened random traces, including 239,035 intermediate and terminal states. It is faster than enumerated screening on METR-LA localized masks but slower on MontevideoBus (Table~\ref{tab:review}). On the METR-LA localized timing sample, packed MOMS takes 4.94 ms, compared with 0.98 ms for component screening. Reducing label queries can therefore require more local computation. The timings exclude graph preprocessing and actual retrieval, so they do not measure communication latency or energy consumption.

\section{Discussion and Conclusion}
The audit determines a specified pattern of interval violations from
partially observed labels. The experiments show that query count, early positive discovery and computation can favor different policies. Node reindexing also changes some comparisons. These findings concern existing selection rules under exact retrieval and simulated visibility masks. Sparse positive timestamps and repeated inspection of the data limit generalization. Exact stopping certifies the defined event, not the statistical coverage of the forecast intervals.

The unit cost model treats every retrieved label equally. Deployment with unequal retrieval costs or parallel requests would require a different acquisition objective. Retrieval errors would also invalidate the exact stopping argument, which assumes that every returned label is correct. The present experiments isolate the choice of query order under these assumptions. They do not establish a policy for noisy or asynchronous measurement systems.
\label{endbody}
\clearpage
\twocolumn
\bibliographystyle{IEEEbib}
\section{Compliance with Ethical Standards}
This study uses only publicly available benchmark datasets and does not
involve the collection of data from human or animal subjects. No ethical
approval was required.

\section{Acknowledgment}
No funding was received for conducting this study. The authors have no
relevant financial or nonfinancial interests to disclose.
\bibliography{references}
\end{document}